%% file: causal_learning.tex
\documentclass{article} 
\usepackage{iclr2020_conference,times}
\input{math_commands.tex}

\usepackage{hyperref}
\usepackage{url}
\usepackage{graphicx}
\usepackage{subcaption}

\title{Causal Learning by a Robot with Semantic-Episodic Memory in an Aesop's Fable Experiment}

\author{Ajaz A. Bhat \\
School of Psychology\\
University of East Anglia\\
Norwich, NR47TJ, UK \\
\texttt{a.bhat@uea.ac.uk} \\
\And
Vishwanathan Mohan \\
School of Computer Science and Electronic Engineering \\
University of Essex \\
Wivenhoe Park Colchester, CO4 3SQ, UK \\
\texttt{vishwanathan.mohan@essex.ac.uk} \\
}

\iclrfinalcopy 
\begin{document}

\maketitle
\begin{abstract}
Corvids, apes, and children solve “The Crow and The Pitcher” task (from Aesop's Fables) indicating a causal understanding of the task. By cumulatively interacting with different objects, how can cognitive agents abstract the underlying cause-effect relations to predict affordances of novel objects? We address this question by re-enacting the Aesop's Fable task on a robot and present a) a brain-guided neural model of semantic-episodic memory; with b) four task-agnostic learning rules that compare expectations from recalled past episodes with the current scenario to progressively extract the hidden causal relations. The ensuing robot behaviours illustrate causal learning; and predictions for novel objects converge to Archimedes' principle, independent of both the objects explored during learning and the order of their cumulative exploration.
\end{abstract}

\section{Introduction}
The ability to learn causal regularities and object affordances allows organisms to exploit objects in the service of their needs and wants, an obvious adaptive advantage in the race for survival. Experiments exploring paradigms like ‘floating peanut’ task \citep{Hanus2011},  trap-tube task \citep{Martin-Ordas2008a} and more recently the Crow and the Pitcher task from Aesop’s Fable \citep{Jelbert2014b, Cheke2012a}  provide evidence that children, primates and corvids can reason to varying degrees about the causally task-relevant properties of objects. A pertinent question therefore is how through cumulative and explorative interactions with different objects in the world, cognitive agents learn task-relevant physical and causal relations and then exploit these flexibly in novel contexts. Accordingly, theoretical works have investigated causal learning in cognitive psychology \citep{Gopnik2004,Griffiths2005}, computer science \citep{Pearl2009,Shimizu2006}, recently in machine learning \citep{ Battaglia2016, Iten2020, Baradel2020} but less so in robotics \citep{Xiong2016}. We explore this question in the context of robotics to show how a semantic-episodic memory system along with a set of four learning rules endows a robot, \textit{iCub} with the capability to cumulatively extract and infer causal relations between objects and actions in "the Crow and the Pitcher" task. In designing this model, we connect some major trends from neuroscience on distributed hub-based semantic memory representation \citep{Kiefer2012,Ralph2017}, small-word properties \citep{Sporns2011}, episodic memory circuitry \citep{Allen2013,Lee2015} to hypothesize possible mechanisms of causal learning. 

\subsection{Causal Learning Task}
\begin{figure}
  \centering
  \includegraphics[trim={0 0.25cm  0 2cm}, width=\linewidth]{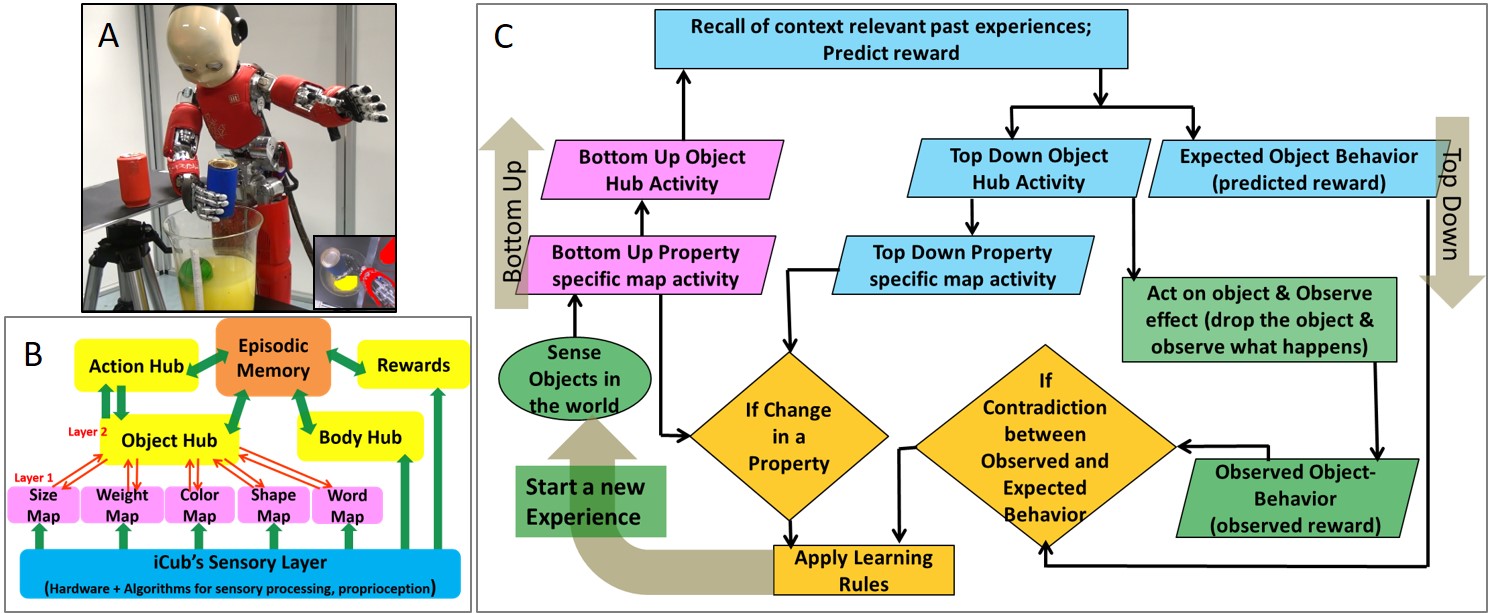}
  \caption{Panel A shows the setup. Each episode involves robot either dropping a single object or making a choice between multiple objects. Objects vary in their physical properties: color, size, shape and weight. Panel B shows the block diagram of the proposed model. Panel C is a flowchart of how the learning rules are applied.}
  \label{figModel}
\end{figure}

The task is inspired from an Aesop’s fable (see analogous empirical works \citep{Jelbert2014b, Bird2009a}), in which a thirsty crow drops pebbles into a half-filled water pitcher, raising the water level high enough to drink. In a comparable scenario (Figure \ref{figModel} A), iCub has a set of objects (on a reachable stand) and a jar of water containing a floating target (green ball) in front. With the goal to reach the green ball (which as such is unreachable), iCub explores if use of available objects can help realize its (otherwise unrealizable) goal. A priori, there is no previous experience with any of the objects. Hence, iCub knows nothing a priori concerning the causal nature of the task. 
\section{Model Description}
\textbf{Hub-based Semantic Memory:} Figure \ref{figModel} B shows the proposed model. At the bottom is the sensory layer to analyze object properties: colour, shape, size and weight. Word labels are provided by the teacher either to issue goals to iCub or teach names of new objects. Output from sensory/physical layer is passed upwards to a bunch of \textit{property-specific} self-organizing maps (SOMs) that encode object properties as concept features. Neural connectivity between the sensory layer and the property-specific maps is learnt using standard SOM procedure (see \citet{Bhat2016d}). There onwards, perceptual analysis of an object (e.g. a \textit{large heavy blue cylinder}) leads to activations in different property-specific SOM’s coding for color, shape, weight and size respectively. In this sense, layer 1 emulates the distributed property-specific organization of perceptual information in brain \citep{Patterson2007, Martin2016}. Activity in these maps forms the bottom up input to the layer 2 SOM i.e. the object hub. The learning rule for dual dyad \citep{Park2013a} connectivity matrix $ W $  from SOMs to the object hub (and its counterpart $ W' $ backwards) is: if the net activity due to a neuron $ i $ and a neuron $ j $ winning in the maps manages to activate a neuron $ k $ in the object hub, set $ W_{ik} = 1$ and $ W_{jk} = 1 $. Thereby, the object hub facilitates integration and multimodal object representation, as evidenced neuro-scientifically \citep{Ralph2017, Kiefer2012}. This small-world network \citep{Sporns2011} of maps and the object hub is complemented with a dynamics to let the neural activity in one map retro-activate other members of the network, hence allowing information to move \textit{top down}, \textit{bottom up} or in \textit{cross modal} fashion. So, a word like “blue cylinder” activates word-map that forwards activations to the object hub, which thereafter retro-activates the shape and colour maps, as if expecting top down what a “blue cylinder” might be perceptually.
\\
In an embodied robot, \emph{objects} in the environment are employed via \emph{actions}. Here, actions are represented at an abstract level (“what can be done with an object”) separated from the action planning details (“how to do”). While the former relates to the motor affordance of an object, the latter relates to motion planning details, and interested reader may refer to \citet{Bhat2017} for details related to motion planning. In layer 2, the abstract representation corresponds to the action hub and consists of single neurons coding for different action goals such as reach, grasp, etc. In this sense, action hub neurons are similar to \textit{canonical} neurons in the pre-motor cortex \citep{Rizzolatti2004} that are activated at the sight of objects to which specific actions are applicable. Finally, the consequences of both perceptions and actions alter the state of the body itself. The body hub, another layer 2 neural map explicitly encodes these states of the body (like failing to reach an object etc.). Reward of an experience is either given by the user or evaluated by the robot itself though observation. In this task, reward is the volume/level of water raised by dropping an object into the jar of water.
\\
\textbf{Episodic memory and its link to the Hubs:} Practically, when a robot interacts with its environment, it is the ongoing sequences of actions on various objects, the ensuing consequences, internal body state, and rewards received that mainly form the content of its experiences. Thus, in our model, it is the temporal sequence of activations in the different hubs (object, action, reward, body) during an episode that make up the episodic memory content. The episodic memory is realized using a excitatory-inhibitory neural network of auto-associative memory \citep{Mohan2014f,Hopfield2008}. It consists of 1000 neurons, organized in a sheet like structure with 20 rows each containing 50 neurons. Every row is an event in time (indicating activation in object hub, action hub, body hub or reward) and the complete memory as an episode of experience. For example, being unable to reach a floating object in a jar of water (Body hub state), perceiving a red cylinder (Object hub state), dropping it in water (Action hub state), fetching a reward of 100 (end reward). In future, if the robot perceives the red cylinder, the object hub state serves as a partial cue to reconstruct the full experience. Importantly, in the memory network of 1000 neurons, multiple episodic memories $ (\approx 230) $ can be represented and retrieved \citep{Hopfield2008}. See \citet{Bhat2016d} for methods to encode new experiences in episodic memory and recall past ones from partial cues.
\\
\textbf{Learning Rules for Causal Abstraction:} As Figure \ref{figModel} C shows, these rules compare \textit{what the robot has experienced in the past} against \textit{what is happening in the present situation}. Let \textbf{$ \bigtriangleup Property $} be the difference in activity in a property-specific map when activated bottom up (through sensory layer) and when activated top down through recall of past event from episodic memory. Let \textbf{$ \bigtriangleup Contradiction $} be the difference between the robots anticipation of how an object might behave (expected reward due to recalled past experience) and the real observed behaviour. Then the learning rules are as follows:
\\
\textbf{(E)limination rule :} If \textbf{$ {\bigtriangleup Property}  \land  {\neg {\bigtriangleup} Contradiction} $}, then that property is not causally dominant and hence drastically reduce the connection strength between the object hub and the associated property-specific map.
\\
\textbf{(G)rowth rule :} If \textbf{$ {\bigtriangleup Property}  \land  {\bigtriangleup Contradiction} $}, then that property is causally dominant. Hence connectivity between the object hub and the associated property-specific is strengthened, and encode the experience in episodic memory.  Contradiction in the robot’s anticipation implies that there is something new to learn.
\\
\textbf{(U)ncertainty rule :} If \textbf{$ {\neg{\bigtriangleup} Property}  \land  {\bigtriangleup Contradiction} $}, the connection strength between the object hub and the map is marginally reduced. In this condition it is not possible to infer whether the property is causally dominant or not, unless further experience is gained.
\\
\textbf{(S)tatus Quo rule :} If \textbf{$ {\neg{\bigtriangleup} Property}  \land  {\neg{\bigtriangleup} Contradiction} $}, nothing new to learn, so no change in connectivity.

\section{Results}
\label{results}
Given that the robot is learning cumulatively, this section presents successive episodes of learning (a video playlist of these experiments is available \href{https://www.youtube.com/playlist?list=PLIfoHEM1gr24EniCzBuUxZ2tqNpQA8QQm}{\emph{online}}). All episodes share a set of computational processes, 1) bottom up experience/interaction with the world and hence activation of various maps 2) recall of past experiences from memory (if any); 3) Use of recalled past experiences to anticipate and 4) application of the learning rules.

\begin{figure}[h!]
  \centering
  \begin{subfigure}[b]{0.6\linewidth}
    \includegraphics[trim={0 0.5cm  0 3cm},width=\linewidth]{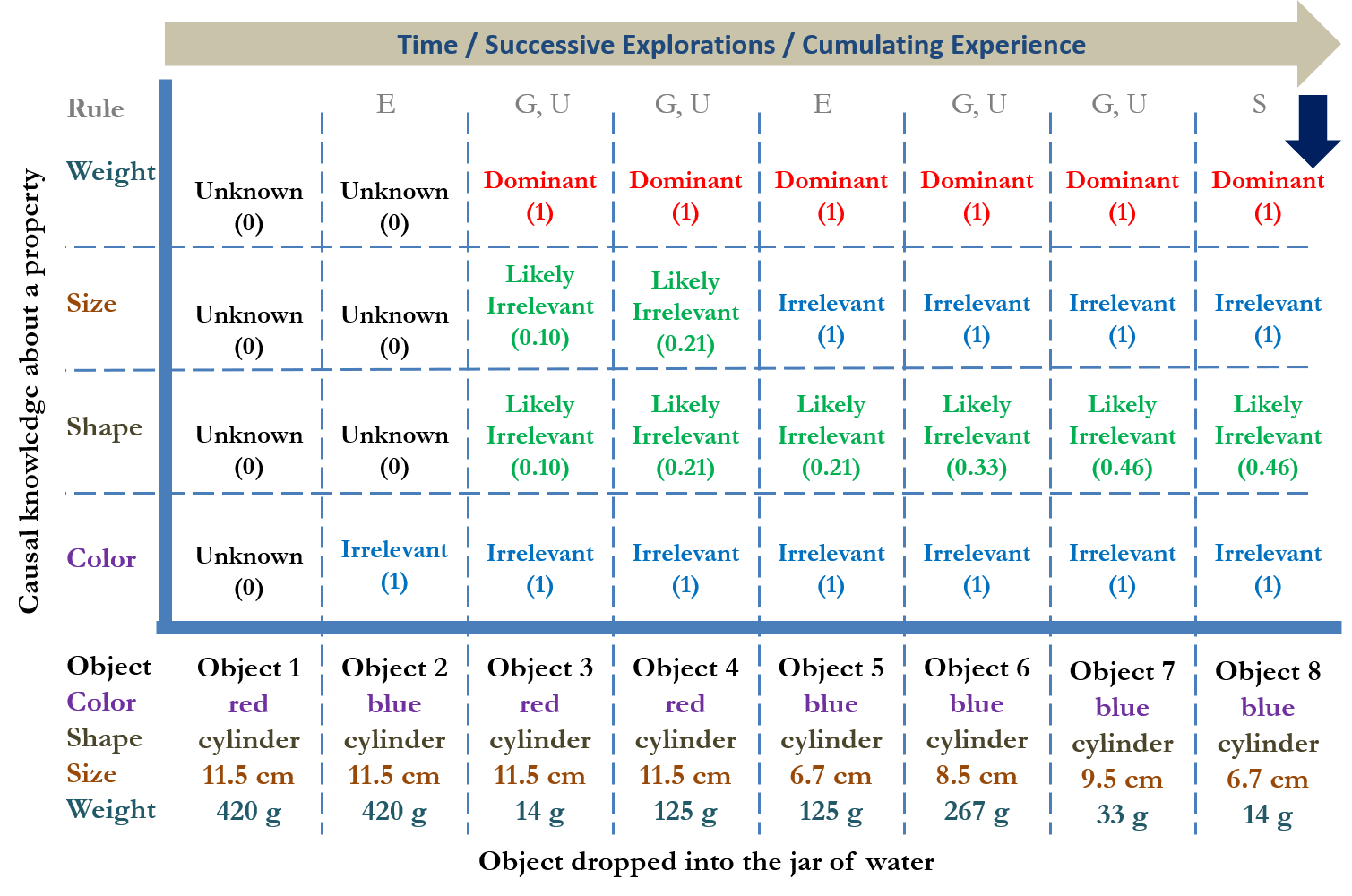}
  \end{subfigure}
  \begin{subfigure}[b]{0.39\linewidth}
    \includegraphics[trim={0 0.5cm  0 3cm},width=\linewidth]{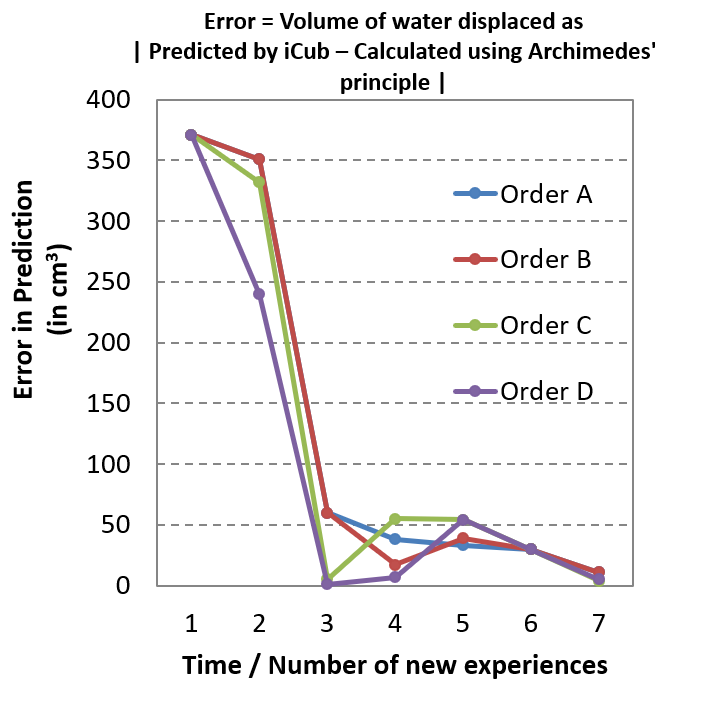}
  \end{subfigure}

  \caption{Left panel plots growing causal knowledge as robot explores objects over successive episodes. Causal knowledge regarding a property is either \textit{Unknown} (depicted by 0 in the plot); or has been learnt to be \textit{Dominant} or \textit{Irrelevant} (depicted by 1); or the system expects the property to be \textit{Likely Irrelevant} ( a certainty value between 0 and 1). Right panel plots error in iCub's prediction about the volume of water displaced on dropping an object against actual volume displacement calculated using Archimedes’ principle. The four curves correspond to four different orders in which the objects were presented to robot for exploration.}
  \label{figResults}
\end{figure}

\textbf{Learning that colour of objects is causally irrelevant to the Aesop’s fable task}
\\
\textit{\textbf{Episode 1:}} In the first episode, iCub is given the goal to reach the green ball (in the jar of water). The motion planning system of iCub provides the information that the goal is unreachable. A \textit{large heavy red cylinder} is available and detected (see Figure \ref{figResults} left, \emph{Object 1} ). Bottom up sensory streams activate property-specific maps related to (red) color, (cylinder) shape, $(11.5 cm)$ size and $(420 g)$ weight properties leading to a distributed representation of the object in the maps and object hub. The object hub activity leads to generation of a partial cue for the recall of any related past episodes. Since there is no previous experience in the episodic memory, nothing is recalled. So, there is no top down activity in the object hub nor any reward expected. With only option to explore,  the robot picks and drops the object into the jar of water. The object sinks in the water displacing a volume of water of about $ 365 cm^{3} $ enough to make the floating green sphere reachable. This experience is encoded into the episodic memory: an unreachable goal (body hub state), dropping a large heavy red cylinder (object hub activity), a volume of water displaced $ 365 cm^{3} $ (reward) and goal realized successfully (body hub state). Note, this is a rapid one-shot event encoding into memory.
\textit{\textbf{Episode 2:}} iCub is presented with a \textit{blue} cylinder of the same size and weight as one in episode 1. Object hub activity (with partial similarity to \emph{Object 1} generates a partial cue that) leads to recall of the only one past experience (i.e. episode 1: Figure \ref {figResults} left, \emph{Object 2}). So iCub anticipates that large heavy \textit{blue} cylinder would displace $ 365 cm^{3} $  of water and this turns out to be the case once the robot actually drops the object into water. Comparing the expected behaviour of object (reward hub activity in the recalled past experience) and the observed behaviour, the robot finds no contradiction. In sum, \textit{\textbf{there is change in a property (colour), but it did not cause any contradiction in the expected behaviour}}. Elimination rule applies here implying colour is not a causally dominant property. The connectivity between the colour map and the object hub is drastically reduced, so they no longer will retro-activate each other. This episode is not encoded into episodic memory.
 
\textbf{Learning that weight is a causally dominant property}
\\
\textit{\textbf{Episode 3:}} iCub is presented with a very \textit{light} cylinder $(14 g)$ with other properties same as in episode 1 (see Figure \ref {figResults} left, \emph{Object 3}) Bottom-up activity recalls episode 1. A high reward of $ 365 cm^{3} $  is anticipated. However, only a small amount $ (24 cm^{3}) $ of water is displaced after robot's action, leading to a contradiction between expected and observed behaviours. A comparison of the bottom up activity and the reconstructed top down activity reveals there is a difference in weight map. Growth rule is applied because \textit{\textbf{a change in the weight property causes contradiction}}. The new experience is encoded into the episodic memory which can be recalled next time for better prediction. Furthermore, activity in shape and size maps showed no change even though there was a contradiction between the expected and the observed behaviour. Hence the Uncertainty rule applies too: as the robot still has no experience or complete knowledge of the causal relevance of object-size or shape. The robot partially believes at this point that ‘shape and size’ of the object may not be relevant in causing water rise. 

Accumulating over a set of such experiences (Figure \ref{figResults} left) with objects of different properties, the robot grows it’s causal knowledge (certainty) of properties relevant to the task. Furthermore, in cases when the system is presented a cylinder of a weight never experienced before, all the past experiences (due to the same shape) will be recalled and a weighted averaging of the rewards expected due to these past experiences is used as an estimate of net anticipated reward (see \citet{Bhat2016d} for more details). As the number of experiences with objects of different weights increases, the accuracy in the prediction of reward increases systematically. Figure \ref{figResults} (right) show robot's predictions in four different random orders of 8 objects. Results show that the causal knowledge is same at the end of explorations in all cases and error in prediction (i.e., difference between robot’s prediction and Archimedes’ principle) rapidly decreases in all cases.

\section{Conclusion}
\label{Conclusion}
The work takes the topic of affordances from the level of \textit{object-action} to the level of \textit{property-action}, in line with emerging studies from neurosciences, and suggests a possible mechanism for causal learning in animals and robots. The work emphasizes that reasoning and learning always have to go hand-in-hand and grow cumulatively and continuously in lifetime of a learner, be it a natural or an artificial cognitive agent.

%

\bibliography{causal_learning}
\bibliographystyle{iclr2020_conference}


\end{document}

%% file: math_commands.tex
\usepackage{amsmath,amsfonts,bm}

\def\eqref#1{equation~\ref{#1}}

\def\1{\bm{1}}

\DeclareMathAlphabet{\mathsfit}{\encodingdefault}{\sfdefault}{m}{sl}
\SetMathAlphabet{\mathsfit}{bold}{\encodingdefault}{\sfdefault}{bx}{n}

